\documentclass[journal,draftcls,onecolumn]{IEEEtran}

\ifCLASSINFOpdf
 
\else
 
\fi

\usepackage{flushend}
\usepackage{csquotes}
\usepackage{graphicx}
\usepackage{amsmath}
\usepackage{amsthm}
\usepackage{amssymb}
\usepackage{color}
\usepackage{algorithm}
\usepackage{algpseudocode}
\usepackage{comment}
\usepackage{multirow}
\usepackage{tabularx}
\usepackage{booktabs}
\usepackage[normalem]{ulem}
\usepackage{cite}
\usepackage{bm}
\useunder{\uline}{\ul}{}
\usepackage{url}
\usepackage{textcomp}
\usepackage{xcolor}
\usepackage{physics}
\usepackage{mathtools}

\usepackage{booktabs}

\definecolor{ggreen}{RGB}{21,138,7}

\newcommand{\bacca}[1]{\textcolor{black}{#1}}
\newcommand{\Hans}[1]{{\textcolor{black}{#1}}}
\newcommand{\Miguel}[1]{{\textcolor{black}{#1}}}
\newcommand{\Ramith}[1]{{\textcolor{black}{#1}}}
\newcommand{\fyp}[1]{\textcolor{ggreen}{#1}}
\graphicspath{{figures/}} 
\newcommand{\edwin}[1]{\textcolor{black}{#1}}
\UseRawInputEncoding

\DeclareMathOperator*{\argmin}{arg\,min}

\theoremstyle{definition}

{
	\theoremstyle{plain}
	
}

\usepackage{pgf,tikz,pgfplots}
\usetikzlibrary{arrows,shapes}
\usetikzlibrary{plotmarks}
\usepgfplotslibrary{groupplots}
\pgfplotsset{compat=newest}
\pgfplotsset{plot coordinates/math parser=false}
\usetikzlibrary{positioning,spy}
\usetikzlibrary{backgrounds}

\usepackage{tcolorbox}
\definecolor{my-blue}{cmyk}{0.1, 0.0, 0.0, 0.00, 0.700}
\definecolor{my-black}{cmyk}{0, 0.0, 0.0, 0.95, 1.00}

    {\endtcolorbox}

\usepackage[caption=false,font=footnotesize]{subfig}
\begin{document}

\title{\huge Deep Optical Coding Design in Computational Imaging }\vspace{-7em}

\author{Henry Arguello\IEEEauthorrefmark{1}\IEEEauthorrefmark{2},
        Jorge Bacca\IEEEauthorrefmark{1}\IEEEauthorrefmark{2},
        Hasindu Kariyawasam\IEEEauthorrefmark{3},
        Edwin Vargas\IEEEauthorrefmark{1},
        Miguel Marquez\IEEEauthorrefmark{1},
        Ramith Hettiarachchi\IEEEauthorrefmark{3},
        Hans Garcia\IEEEauthorrefmark{1},
        Kithmini Herath\IEEEauthorrefmark{3},
        Udith Haputhanthri\IEEEauthorrefmark{3},
        Balpreet Singh Ahluwalia\IEEEauthorrefmark{4},
        Peter So\IEEEauthorrefmark{5}, 
        Dushan N. Wadduwage\IEEEauthorrefmark{6},
        Chamira U. S. Edussooriya\IEEEauthorrefmark{3} \\
        \IEEEauthorrefmark{1}Universidad Industrial de Santander, Bucaramanga, Colombia \\
        \IEEEauthorrefmark{3}Department of Electronic and Telecommunication Engineering, University of Moratuwa, Sri Lanka\\
        \IEEEauthorrefmark{4}Department of Physics \& Technology, UiT The Arctic University of Norway, Norway. \\
        \IEEEauthorrefmark{5}Department of Mechanical and Biological Engineering, Massachusetts Institute of Technology, USA \\
        \IEEEauthorrefmark{6}Harvard's Center for Advanced Imaging, Harvard University, USA \\
        \IEEEauthorrefmark{2}All these authors contributed equally. \vspace{-2em} \thanks{This work was partially supported by the Air Force Office of Scientific Research under award number FA9550-21-1-0326, the Research Council of Norway, INTPART project (No: 309802), NIH grants R21-MH130067 and P41-EB015871, the Center for Advanced Imaging at Harvard University, the John Harvard Distinguished Science Fellowship Program, and Fujikura Inc. } }
 \vspace{-2em}
\markboth{IEEE Signal Processing Magazine}%
{Shell \MakeLowercase{\textit{et al.}}: Bare Demo of IEEEtran.cls for IEEE Journals}
 \vspace{-2em}
\maketitle
 \vspace{-2em}

\begin{abstract}
Computational optical imaging (COI) systems leverage optical coding elements (CE) in their setups to encode a high-dimensional scene in a single or multiple snapshots and decode it by using computational algorithms. The performance of COI systems highly depends on the design of its main components: the CE pattern and the computational method used to perform a given task. Conventional approaches rely on random patterns or analytical designs to set the distribution of the CE. However, the available data and algorithm capabilities of deep neural networks (DNNs) have opened a new horizon in CE data-driven designs that  jointly consider the optical encoder and computational decoder. Specifically, by modeling the COI measurements through a fully differentiable image formation model that considers the physics-based propagation of light and its interaction with the CEs, the parameters that define the CE and the computational decoder can be optimized in an end-to-end (E2E) manner. Moreover, by optimizing just CEs in the same framework, inference tasks can be performed from pure optics.  This work surveys the recent advances on CE data-driven design and provides guidelines on how to parametrize different optical elements to include them in the E2E framework. Since the E2E framework can handle different inference applications by changing the loss function and the DNN, we present low-level tasks such as spectral imaging reconstruction or high-level tasks such as pose estimation with privacy preserving enhanced by using optimal task-based optical architectures. Finally, we illustrate classification and 3D object recognition applications performed at the speed of the light using all-optics~DNN.


\end{abstract}

\IEEEpeerreviewmaketitle

 \vspace{-1em}
\section{Introduction}
\IEEEPARstart{W}e are in a new era of computational optical imaging (COI) systems that break traditional sensing limits leveraged by computational signal processing.  COI systems have enabled the acquisition of high-dimensional information of the scene such as polarization~\cite{fu2015compressive}, spectrum~\cite{arce2014compressive}, depth~\cite{sitzmann2018end}, time~\cite{iliadis2020deepbinarymask} with  unprecedented performance in commercial applications such as medical imaging, precision agriculture, surveillance, and vision-guided vehicles \cite{mait2018computational}. With current measurement devices it is only possible to acquire low-dimensional intensity values of high-dimensional scenes. Therefore, the main goal of computational imaging design is to encode the desired information in a low-dimensional sensor image by introducing optical coding elements (CEs) in the optical setups.
The most popular CEs include coded apertures (CAs) ~\cite{arce2014compressive}, constructed by an arrangement of opaque or blocking elements that spatially modulate the wavefront; diffractive optical elements (DOEs)  ~\cite{sitzmann2018end}, which are phase relief elements that use micro-structures to alter the phase of the light propagated through them; color filter arrays (CFAs) ~\cite{chakrabarti2016learning}, which are mosaics of tiny band-pass filters placed over the pixel sensors (or image plane) to capture wavelength information.	
 The decoding or recovery procedure of a high-dimensional image from the coded measurements is carried out in a post-processing step leveraging computational algorithms. The reconstruction quality mainly depends on  the CE pattern and the image processing algorithm employed to decode the desired information. Thus, several works have covered these two fundamental lines of research. Over the last years, different CE designs have been proposed, adopting random patterns, hand-crafted assumptions~\cite{hinojosa2018coded}, or theoretical constraints such as the mutual coherence~\cite{arce2014compressive} or the concentration of measure~\cite{arce2014compressive}. 
On the other hand, alternative studies propose computational processing methods to perform inference tasks directly from the coded measurements avoiding the reconstruction step. Specifically, in \cite{davenport2010signal}, theoretical and simulation results showed that it is possible to learn features directly from the coded measurements. These features are used as input of classifiers such as support vector machines \cite{calderbank2012finding}, sparse subspace clustering \cite{hinojosa2018coded}, and recently, \textcolor{black}{in deep models that are applied for different tasks}~\cite{bacca2021deep}. The unprecedented gains in performance of deep neural networks (DNNs) have rapidly converted them into a standard algorithmic approach to process the coded measurements.
\begin{figure}[!t]
	\centering
	\includegraphics[width=0.95\textwidth]{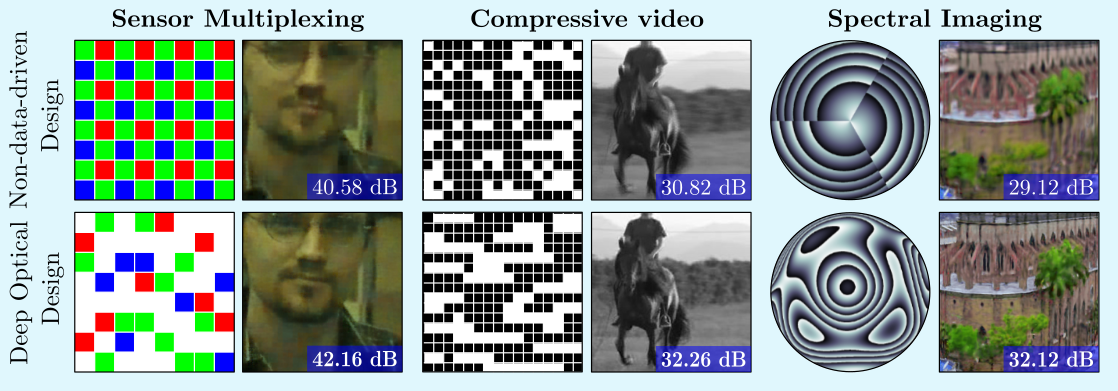}\vspace{-1em}
	\caption{Comparison of non-data-driven CEs to deep optical CEs design  \textcolor{black}{in terms of PNSR for} applications of sensor multiplexing (first column), Compressive Video (second column) and Spectral Imaging (third column). The CEs correspond to a color filter array~\cite{chakrabarti2016learning}, a binary-coded aperture~\cite{iliadis2020deepbinarymask}, and a DOE~\cite{arguello2021shift}, respectively.\vspace{-2em}}
	\label{fig:visual}
\end{figure}
The continuous progress in deep learning and the growing amount of image data have enabled new data-driven CE design where the optical encoder and computational decoder are jointly designed. More precisely, it consists of simulating the COI system as a fully differentiable image formation model that considers the physics-based propagation of light and its interaction with the CEs. In this model, the CE can be represented by learnable parameters and can be interpreted as an optical layer. In the same way, the overall COI system can be interpreted as an optical encoder composed of different optical layers. The optical encoder can be coupled with a DNN or differentiable algorithm in a deep learning model where the ensemble parameters (CEs and DNN parameters) can be optimized in an end-to-end (E2E) manner using a training dataset and a back-propagation algorithm. The optimized CE \textcolor{black}{provides} novel data-driven optical designs that outperform the conventional non-data-driven approaches \textcolor{black}{(see Fig.~\ref{fig:visual} for some visual examples or refer to  \cite{arguello2021shift,bacca2020coupled,bacca2021deep,iliadis2020deepbinarymask,chakrabarti2016learning} for more details).} Interestingly, it has also been shown that a cascade of multiple optical layers can be trained to perform a specific task such as classification \cite{Lin2018}, image-formation \cite{Lin2018}, 3D object recognition \cite{Shi2021MultipleviewRecognition}, or saliency segmentation \cite{Yan2019Fourier-spaceNetwork}. This all-optical deep learning framework is referred to as diffractive deep neural networks (D2NNs). Each optical layer represents a CE, and the transmission coefficient at each spatial location is treated as a learnable parameter. Then the optical model can be trained in an E2E manner to perform a specific task. In this context, the optical system also works as an inference algorithm, and the inference can be achieved from pure optics at the speed of light.

An important aspect of the E2E scheme is the optical system's constraints that must be considered in the coupled deep training model; for example, the lens surface must be regular and smooth, or the attenuation values in the coded aperture could be binary. Additionally, some critical assembling properties and modeling considerations such as the amount of light, and the number of projections, must be considered in the learning-based design of the optical system. These constraints have been tackled in the modeling by parameterizing of the CE or the optimization problem through regularization functions~\cite{bacca2021deep}. After finding the CE optimal parameters of the COI system and the decoder, the optical elements are carefully manufactured or implemented in an optical setup that will acquire the optimized coded measurements. The optimal COI measurements are fed to the trained decoder to perform the inference task of interest. In some cases, the mismatch of the simulated forward model and the transfer function of the experimental setup can demand an additional optical setup calibration, and also a neural network retraining (or fine-tuning)~\cite{sitzmann2018end}. 

This article surveys the recent advances and foundations on data-driven CE design. Our goal is to provide readers with step-by-step guidance on how to efficiently model optical systems and parametrize different CE for being included in the E2E optimization framework. Furthermore, we describe how to address some physical constraints or assembling properties appealing to a wide range of COI systems. After providing a tutorial on how to include the COI system in a deep model, we demonstrate the flexibility of the deep optical design in several real applications including the implementation of D2NNs.
Finally, we discuss current challenges and possible
future research directions.

\section{Computational Imaging System based on Coding Elements}
\label{section:2}
	\begin{table}[!t]
        \centering
        \caption{COI systems summary: coding optical element, encoding dimension, and the design parameters. \textcolor{black}{(Only some references are cited, for more details please see~\cite{cao2016computational,yuan2021snapshot,mait2018computational,wetzstein2020inference} )}}\vspace{-1em}
        \begin{tabular}{|l|l|l|c|} 
        \toprule
        \textbf{Customizable optical element}                                                                             & \multicolumn{1}{c|}{\begin{tabular}[c]{@{}c@{}}\textbf{Encoded} \\ \textbf{dimensions}\end{tabular}} & \begin{tabular}[c]{@{}l@{}}\textbf{Design} \\ \textbf{parameters}\end{tabular}            & \textbf{Ref.}                     \\ 
        \midrule
        \begin{tabular}[c]{@{}l@{}}\textbf{Binary coded aperture}~\textcolor{black}{is constructed by an} \\\textcolor{black}{arrangement of opaque or blocking elements that} \\\textcolor{black}{spatially modulate the wavefront.}\end{tabular}                                                                                                      & \begin{tabular}[c]{@{}l@{}}Holography\\ Spectral\\ Angular\\ Depth\end{tabular}                      & \begin{tabular}[c]{@{}l@{}}Spatial \\ distribution\end{tabular}                           &                        \cite{iliadis2020deepbinarymask,liang2022deep}                \\ 
        \hline
        \begin{tabular}[b]{@{}l@{}}\textbf{Color filter array} \textcolor{black}{ is constructed using a mosaic } \\ \textcolor{black}{of tiny band-pass filters placed over the sensor pixels} \\ \textcolor{black}{(or image plane) to encode wavelength information.}\end{tabular}                                                                                                     &  \begin{tabular}[c]{@{}l@{}}Spectral \\ \textcolor{white}{.} \end{tabular}          & \begin{tabular}[b]{@{}l@{}} Band pass \\ filter\\ distribution\end{tabular}        & \begin{tabular}[c]{@{}l@{}}\cite{arguello2021shift,chakrabarti2016learning} \\ \textcolor{white}{.} \end{tabular}  \\
        \hline
        \begin{tabular}[c]{@{}l@{}}\textbf{Liquid crystal on silicon~} \textcolor{black}{is constructed with a} \\\textcolor{black}{liquid crystal layer sandwiched between a thin-film} \\\textcolor{black}{transistor and a silicon semiconductor to modulate light.}\end{tabular} & \begin{tabular}[c]{@{}l@{}}Spectral\\ Angular\end{tabular}                                           & \begin{tabular}[c]{@{}l@{}}Height \\ map\end{tabular}                                     &                               \cite{hinojosa2021learning,bacca2019super}          \\ 
        \hline
        \begin{tabular}[c]{@{}l@{}}\textbf{Diffractive optical element} \textcolor{black}{is constructed by using }  \\ \textcolor{black}{ micro-structures  to alter  the  phase  of the light} \\ \textcolor{black}{propagated through them.}\end{tabular}                                             & \begin{tabular}[c]{@{}l@{}}Spectral\\ Depth\end{tabular}                                             & \begin{tabular}[c]{@{}l@{}}Height \\ map\end{tabular}                                     &                           \cite{sitzmann2018end,arguello2021shift}              \\ 
        \hline
        \begin{tabular}[c]{@{}l@{}}\textbf{Structured illumination~} \textcolor{black}{is based on the projection}  \\ \textcolor{black}{ of patterns onto the object (active illumination), } \\ \textcolor{black}{ which is then viewed by a sensor.}\end{tabular}                                                                                                                               & Depth                                                                             & \begin{tabular}[c]{@{}l@{}}Pattern \\ distribution \\ and intensity\\ values\end{tabular} &                      \cite{bell1999structured,kellman2019physics}                   \\ 
        \hline
        \begin{tabular}[c]{@{}l@{}}\textbf{Deformable mirror~} \textcolor{black}{is constructed by a set of  micro}  \\ \textcolor{black}{mirrors whose surface can be deformed to achieve}  \\ \textcolor{black}{wavefront control and correction of optical aberrations.}\end{tabular}                                                                                                                                                                          & \begin{tabular}[c]{@{}l@{}}Spectral\\ Depth\end{tabular}                                             & \begin{tabular}[c]{@{}l@{}}Height \\ map\end{tabular}                                     &                                       \cite{marquez2019compressive,marquez2021snapshot}  \\
        \bottomrule
        \end{tabular}
        \label{Tb:1}\vspace{-2em}
\end{table} 
%
%
One of the main challenges in computational imaging is to acquire high-dimensional scenes into a low-dimensional intensity detector. Researchers have coincided with modeling the high-dimensional world information with up to eight-dimensional function $f(x,y,z,\alpha,\psi,t,\lambda,p)$, called the plenoptic function, where $(x,y)$ stand for spatial, $z$ for depth, $(\alpha,\psi)$ for angular views, $t$ for temporal, $\lambda$ for wavelength, and $p$ for polarization dimension~\cite{mait2018computational}. \textcolor{black}{In this path, several authors have proposed the design of sophisticated imagers based on CEs to sense two or more dimensions of the plenoptic function (see \cite{cao2016computational,yuan2021snapshot} for more details).} Table~\ref{Tb:1} summarizes some of the optical systems that employ CEs in their optical setups to encode, and then, recover some wavefront dimensions. In the following, we present two fundamental steps for obtaining a physics-based model of the COI system that will be integrated later into the deep optical design. The first step defines the more suitable light propagation model, which is selected based on the light source conditions, the propagation distances, the optical elements, and the target physical constraints.

The Rayleigh-Sommerfeld (RS) diffraction formulation is the scalar diffraction model that allows obtaining a suitable solution for the output field of a given input field under some physical conditions~\cite{Lin2018}. Specifically, considering an input field $U^{0}$ (usually given by the scene of interest), the resultant field of the diffracted optical wave at a given spatial point is denoted as
\begin{equation}
    \label{eq:rayleigh_sommerfeld}
    U^{1}(x_1,y_1,z_1;\lambda) = \iint\limits_A U^{0}(x_0,y_0,z_0;\lambda)\,\left(\frac{z_1 - z_0}{r^2}\right) \left(\frac{1}{2\pi r} + \frac{1}{j \lambda}\right) \exp \left(\frac{j 2\pi r}{\lambda} \right)\,dA,
\end{equation}
where $A$ is the area of the input field.
Here, $r = \sqrt{(x_1 - x_0)^2 + (y_1 - y_0)^2 + (z_1 - z_0)^2}$ is the distance between the spatial points, $\lambda$ is the wavelength of the light and $j = \sqrt{-1}$. \bacca{Although the RS offers high precision propagation modeling, the majority of COI systems employ the angular spectrum~\cite{Lin2018},  Fresnel~\cite{arguello2021shift} or Fraunhofer~\cite{arce2014compressive} diffraction approximation models for practicality, since they are accurate enough.} For instance, the angular spectrum (AS) method is an efficient approach to approximate the RS propagation model. It uses the Fourier relationship between the electric/magnetic field of a light wave and its angular spectrum to perform the computations in the Fourier domain. Here, the angular spectrum of the field is propagated using the transfer function given by,
\begin{equation}
    \label{eq:prop_tf}
    P(z) = \exp\left(j2\pi z\,\sqrt{\frac{1}{\lambda^2} - f_x^2 - f_y^2}\right),
\end{equation}
where $f_x$ and $f_y$ are the two-dimensional (2D) spatial frequency components~\cite{Lin2018}.

\begin{figure}[!t]
\centering
\includegraphics[width=0.93\textwidth]{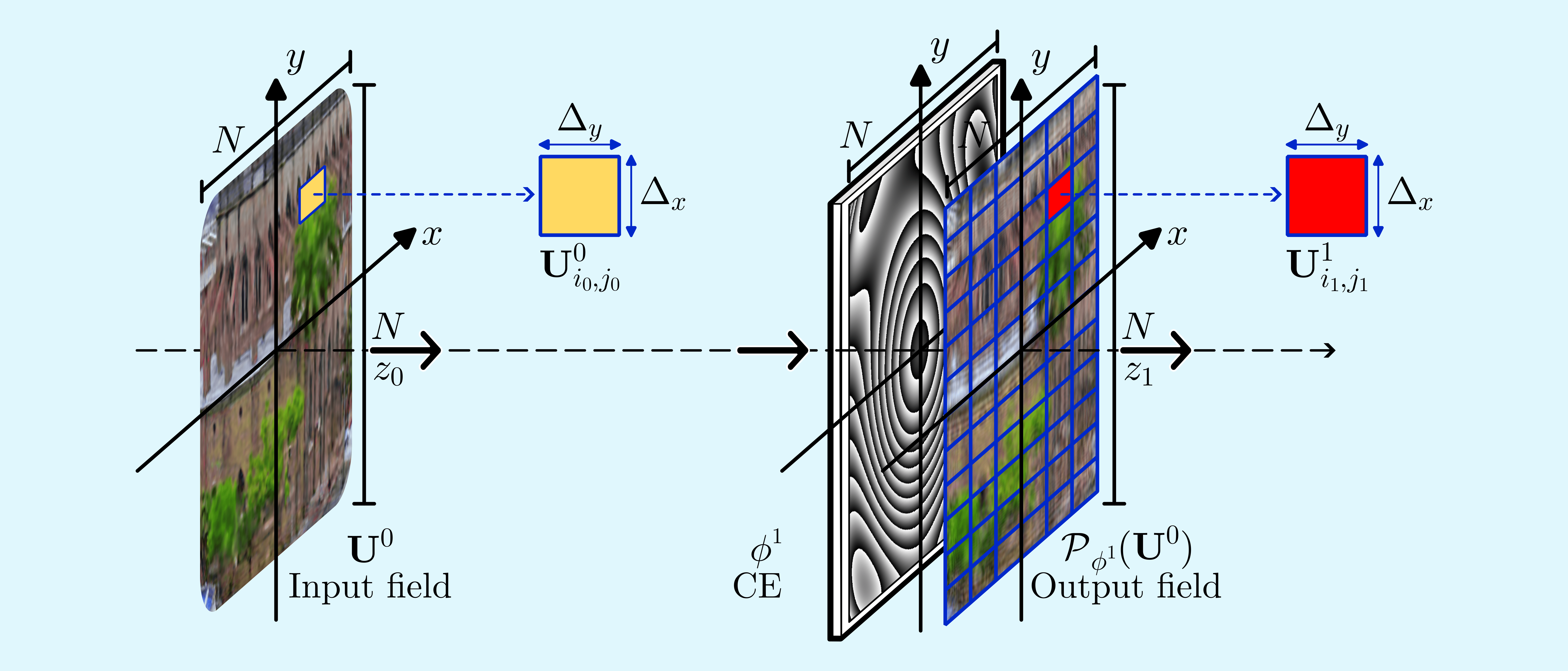}\vspace{-1em}
\caption{Visual representation of the discretized propagation and the modulation of a CE in an optical layer. The incident field $\mathbf{U}_0$ is propagated and then modulated by the CE $(\boldsymbol{\phi})$ where the field immediately after the $i_1,j_1$ pixel (or neuron) is given by  $\mathbf{U}^1_{i_1,j_1}$.} 
\label{fig:d2nn_modelling}\vspace{-2em}
\end{figure}

Once the light propagation model is selected, the following step consists of modeling the interactions with the CE. According to the CE physical composition and structure, the incoming wavefront is modulated in its phase~\cite{bacca2019super} or intensity~\cite{iliadis2020deepbinarymask}. The current fabrication technologies limit the CE structures to discrete distributions. For that reason, we restrict our attention to work with the propagation and modulation model in a discrete form. For instance, let the DOE be considered as a CE. The propagation of a discrete input field $\mathbf{U}^0$ at plane $z_0$ and its interaction with the discretized DOE with $N\times N$ pixels located in plane $z_1$ is illustrated in Fig.~\ref{fig:d2nn_modelling}. Each DOE's pixel has a particular light response, known as the transmission coefficient. Based on Riemann summation of the integral \eqref{eq:rayleigh_sommerfeld}, the propagation of $\mathbf{U}^0$ up to the DOE as well as the DOE's modulation can be mathematically expressed as 
\begin{equation}
    \label{eq:direct1}
    \mathbf{U}^1_{i_1,j_1} =  \boldsymbol{\phi}_{i_1,j_1}^{1}\cdot\sum_{i_{0}=0}^{N-1}\sum_{j_{0}=0}^{N-1} \mathbf{U}^0_{i_0,j_0}\,w\,\Delta_x \Delta_y \coloneqq \mathcal{P}_{\phi^1}(\mathbf{U}^0) , 
\end{equation}
where
\begin{equation}
    \label{eq:direct2}
    w = \left(\frac{z_1 - z_{0}}{r^2}\right) \left(\frac{1}{2\pi r} + \frac{1}{j \lambda}\right) \exp \left(\frac{j 2\pi r}{\lambda} \right),
\end{equation}
with $r = \sqrt{(\Delta_x i_{1} - \Delta_x i_0)^2 + (\Delta_y j_1 - \Delta_y j_0)^2 + (z_{1} - z_{0})^2}$, $\{\Delta_x,\Delta_y\}$ are the width and height pixel resolution, $\mathbf{U}^{0}_{i_0,j_0}$ is the input diffracted wavefront at pixel $(i_0,j_0)$, $\mathbf{U}^{1}_{j_1,j_1}$ is the modulated field at the spatial point $(i_1,j_1)$ just after the DOE, and $\boldsymbol{\phi}_{i_1,j_1}^{1}= \mathbf{a}^{1}_{i_1,j_1} e^{j \boldsymbol{\psi}^{1}_{i_1,j_1}}$ is the transmission coefficient at position $(i_1,j_1)$ with $\mathbf{a}^{1}_{i_1,j_1}$ and $\boldsymbol{\psi}^{1}_{i_1,j_1}$ being the amplitude and the phase terms, respectively. In this case, light modulation is performed by the complex transmission coefficient of the DOE, which is the primary customized variable. The operator $\mathcal{P}_{\boldsymbol{\phi}^1}(\cdot)$ represents the light propagation and modulation of the CE $(\boldsymbol{\phi}^1)$ which is referred to as the optical layer in the E2E framework.\footnote{Notice that $\mathcal{P}_{\boldsymbol{\phi}^1}(\cdot)$ also depends on the destination coordinates $i_1,j_1,z_1$ that we omit in the notation for simplicity.} 

Usually, COI systems comprise more than one CE, as is the case of the D2NN, which is composed of $n$ sequential concatenated CEs (optical layers). Consequently, the encoded wavefront in \eqref{eq:direct1} continues propagating for a set of additional arbitrary CEs to finally converge to the sensor, which converts photons to electrons, i.e., it measures the photon flux per unit of surface area. A general sensing process can be expressed as 
\begin{equation}
    \mathbf{G} = \mathcal{S}\left( \mathcal{P}_{\boldsymbol{\phi}^n}(\mathcal{P}_{\boldsymbol{\phi}^{n-1}}(\cdots \mathcal{P}_{\boldsymbol{\phi}^{1}}(\mathbf{U}^0)\cdots))\right), 
    \label{eq:propagations}
\end{equation}
where $ \mathcal{S}(\cdot)$ denotes the transfer function of the sensor and $\mathbf{G} \in \mathbb{R}^{N \times N}$ is the detected intensity, \textcolor{black}{known as coded measurements.}
For convenience of notation, we will henceforth represent the COI system in vector form as
\begin{equation}
\label{eq:}
    \mathbf{g}=\mathcal{M}_{\boldsymbol{\phi}}(\mathbf{f}),
\end{equation}
where $\mathbf{g}\in \mathbb{R}^{m}$ denotes the coded measurement which is the vector representation of $\mathbf{G}$, \textcolor{black}{$\mathbf{f} \in \mathbb{C}^n$ (or $\mathbb{R}^n$)} represents the underlying scene given by the input field, and $\mathcal{M}_{\boldsymbol{\phi}}$ denotes the mapping function that models the three basic COI systems operators: propagation, interaction with CEs $\boldsymbol{\phi} = \left\{\boldsymbol{\phi}^l\right\}_{l=1,2,\dots,n}$, and sensing process. Although the modeling expressed in \eqref{eq:propagations} is accurate, depending on the application, the operator $\mathcal{M}_{\boldsymbol{\phi}}$ can be further simplified. For imaging applications, where we consider that different points of a scene are added incoherently, the operator can be modeled by a shift invariant convolution of the image (intensity) and a point spread function (PSF) defined by ${\boldsymbol{\phi}}$. Another simplification is employed in a CA-based system where the modulation is just in amplitude. In this case, the COI system can be modeled as a linear system, and thus the operator $\mathcal{M}_{\boldsymbol{\phi}}$ can be expressed as a sensing matrix whose values depend on the CA. In the next section, we present the deep optical coding design framework that requires the appropriate COI sensing model.


\section{Deep Optical Coding Design}

\begin{figure}[!t]
	\centering
	\includegraphics[width=0.95\textwidth]{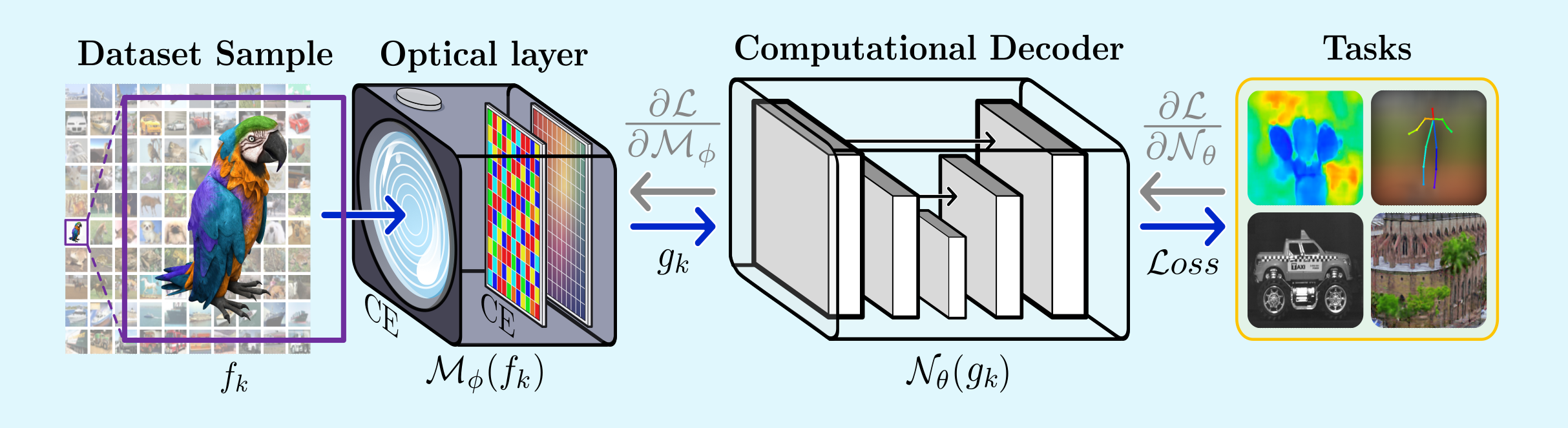}\vspace{-1em}
	\caption{E2E scheme where the COI system is modeled as optical layers. In training, a set of  images passes through the optical system, obtaining the projected measurements that enter the computational decoder producing an output for an arbitrary task.  The estimated task error is propagated from the output of the decoder to the optical layer, updating the weights of the decoder and the optical system.} \vspace{-1em}
	\label{fig:E2E_scheme}
\end{figure}

\textcolor{black}{Optical coding design based on deep learning has shown considerable improvement for multiple
computational imaging applications compared with analytical or mathematical optical coding design.} Specifically, the deep optical coding design jointly optimizes the CEs of a COI system and the computational algorithm or DNN for a specific task in an E2E manner. The  key  idea for  an  E2E  design  is  to  accurately simulate the COI measurements, as is explained in section~\ref{section:2}, 
and to employ them as an input of a differentiable computational algorithm that performs a given task. Therefore, the task error, defined by a loss function, can be propagated to the trainable parameters of the algorithm and further to the COI system parameters to update them towards an optimal point.
In this sense, we can consider the model of the COI system and the DNN as a coupled DNN (see Fig.~\ref{fig:E2E_scheme}), where the physical optical layers can be interpreted as an optical encoder whose learnable parameters describe the CEs.
\subsection{Optical Encoder}
\begin{figure}[!t]
	\centering
	\includegraphics[width=1\textwidth]{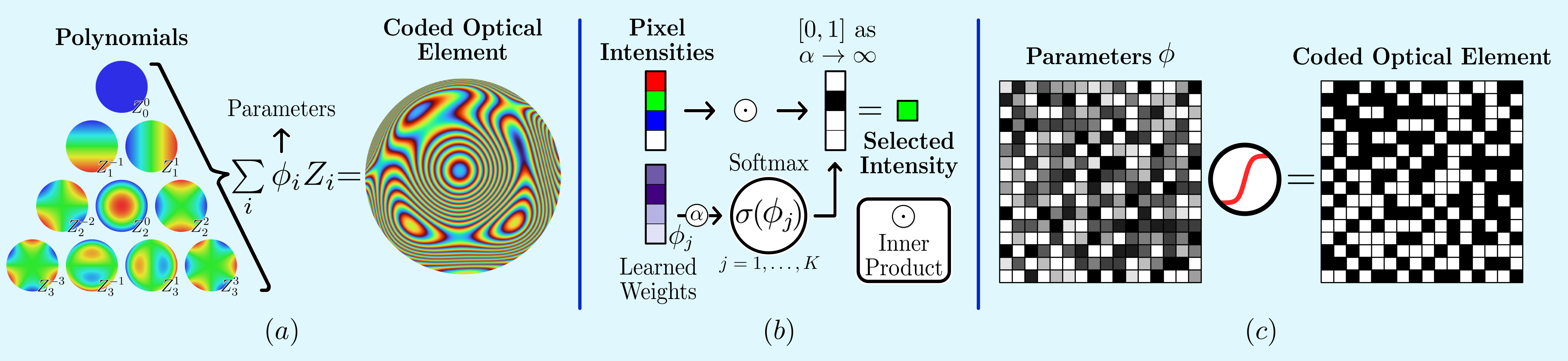}\vspace{-1em}
	\caption{Parametrization of different optical elements, (a) DOE, where the trainable parameters are the coefficients associated with Zernike polynomial $Z_i$ (b) color filters, where the parameters are a vector of weights that select one of the available filters, and (c) coded aperture, where the trainable parameter is a continuous variable. }
	\label{fig:Parametrization}\vspace{-2em}
\end{figure}

\textcolor{black}{To model the real physical system, the following are necessary: an accurate propagation of light, a correct parametrization of the CEs and a proper sensing model.}  Leveraging auto differentiation tools to efficiently optimize neural networks, it is only necessary to employ a differentiable model of $\mathcal{M}_{\boldsymbol{\phi}}(\cdot)$ to efficiently optimize the CEs $\boldsymbol{\phi}$ using a backpropagation algorithm. Therefore, the key point is the CE parametrization that accounts for a simple and accurate representation of the optical element. For instance, the height map of the DOE needs to be regular and smooth for its easy fabrication. Thus, the work in~\cite{sitzmann2018end} proposed to use the Zernike polynomial to describe the height map of the DOE; consequently, the CE parameters $\boldsymbol{\phi}$ are the coefficients of the Zernike polynomials as expressed in Fig(\ref{fig:Parametrization})(a), i.e., instead of learning the height of each pixel, it is only necessary to learn the coefficient associated with each Zernike polynomial.  Another example includes the multiplexing pattern design of a sensor array to reconstruct an RGB image from a single 2-dimensional sensor~\cite{chakrabarti2016learning}. In this case, the set of parameters $\boldsymbol{\phi}$ corresponds to the multiplexing pattern that selects which color must be assigned to each \Hans{pixel on the} sensor  as illustrated in Fig(\ref{fig:Parametrization})(b). To model $\boldsymbol{\phi}$ as a selector, it 
is parameterized using either the soft-max function \cite{chakrabarti2016learning} or the sigmoid function \cite{iliadis2020deepbinarymask} to guarantee binary values as illustrated in Fig(\ref{fig:Parametrization})(c). In some scenarios, the parametrization of the optical elements consists in \Hans{using} non-differentiable functions such as the step function~\cite{chakrabarti2016learning} employed in discrete values or a thresholding operation~\cite{iliadis2020deepbinarymask} used in binary optimization. The gradient needs to be approximated or modified to achieve the desired direction in these scenarios. A clear example is when the sign function is used to binarize; in this case, authors in~\cite{iliadis2020deepbinarymask} define the gradient of that function with the identity. \vspace{-1em}

\subsection{Computational Decoder}

Depending on the task to be performed, different decoders have been employed in the E2E framework \cite{bacca2021deep}. For instance,\cite{sitzmann2018end} develops a fully differentiable Wiener filtering operator to recover a deblurred and super-resolution image. On the other hand, the remarkable advances of DNNs have excelled in various tasks. For example, auto-encoders, residual-networks, or the well-known UNet, have been used for image recovery. Furthermore, unrolled-based approaches exploit iterative reconstruction algorithms as layers \cite{jacome2022d}. More recently, DNN with self-attention mechanisms along spatial or spectral dimensions have also been proposed \cite{cai2021mask}. \Hans{Furthermore}, complex architectures to perform high-level tasks such as classification or human pose estimation \cite{hinojosa2021learning} have been employed. 
In some cases, the coded \textcolor{black}{measurements} do not necessarily share the same spatial resolution of the target image. For example, in the coded aperture snapshot spectral imagers (CASSI) \cite{arce2014compressive}, where a dispersive element is used, the coded optical \textcolor{black}{measurements are} spatially distorted. This spatial mismatch prevents some state-of-the-art deep models to be directly applied since they require the spatial resolution of the input and output images to match. Consequently, some lifting strategies are included as an intermediate step between the optical layer and the neural network decoder to obtain the appropriate image size. Commonly, in linear system,s of the form $\mathbf{g=}\mathcal{M}_{\boldsymbol{\phi}}(\mathbf{f})\mathbf{=Hf}$, the lifting of the \textcolor{black}{measurements} can be performed via the transpose operator $\mathbf{H}^T\mathbf{g}$\cite{bacca2020coupled} or the pseudoinverse $\mathbf{H}^\dagger \mathbf{g}$\cite{bacca2021deep}. Another suitable alternative is to add some layers for learning the transpose operator\cite{hinojosa2021learning}. 

\subsection{End-to-End Optimization}
Once the COI system ($\mathcal{M}_{\boldsymbol{\phi}}$) and the decoder ($\mathcal{N}_{\theta}$) have been modeled as layers, the E2E optimization consists of training the encoder-decoder parameters \Hans{using the following optimization problem} 
\begin{equation}
\label{eq:E2E_general}
\hspace{-0em}\{\boldsymbol{\phi}^*,\boldsymbol{\theta}^*\} \hspace{-0em}=\hspace{-0em}\argmin_{\boldsymbol{\phi}, \boldsymbol{\theta}} \textcolor{black}{\mathcal{L}(\boldsymbol{\phi,\theta}):=} \sum_{k}\hspace{-0em} \mathcal{L}_
		{task}\left(\mathcal{N}_{\boldsymbol{\theta}}\left(\mathcal{M}_{\boldsymbol{\phi}}(\mathbf{f}_k )\right) , \mathbf{d}_k\right)\hspace{-0em} +\hspace{-0em}\rho R_\rho(\boldsymbol{\phi}) +\sigma R_\sigma(\boldsymbol{\theta}),
	\end{equation}
where $\{\boldsymbol{\phi}^*,\boldsymbol{\theta}^*\}$ represent the set of optimal optical coding parameters and the optimal weights of the network, respectively,  $\{\mathbf{f}_{k},\mathbf{d}_k\}_{k = 1}^{K}$, 
{account for} the training database with $K$ elements, with $\mathbf{f}_{k}$ as the input image and $\mathbf{d}_{k}$ as the output of the neural decoder that can be a target image, a classification vector, a segmentation map, among others. The loss function $\mathcal{L}_{task}$ is linked to a specific {inference} task. For instance, the mean-squared error (MSE) and cross-entropy metrics are {conventionally} used for reconstruction and classification tasks, respectively~\cite{bacca2021deep}.  $R_\rho(\boldsymbol{\phi})$ and $ R_\sigma(\boldsymbol{\theta})$ denote regularization functions that act in the optical parameters and the weights of the decoder, respectively, with  $\rho$ and $\sigma$ as regularization parameters. The regularization functions have been widely used for training neural networks, since $ R_\sigma(\boldsymbol{\theta})$ has shown to reduce the overfitting problem,  a common issue that appears when training deep neural networks. For instance, the $||\boldsymbol{\theta}||_2$ or $||\boldsymbol{\theta}||_1$ has been successfully applied~\cite{bacca2021deep}.

\begin{table}[t!]
     \begin{center}
           \caption{Examples of the regularization on the coded apertures to obtain specific characteristics}\vspace{-1em}
      \label{tbl:Regularizations}
     \begin{tabular}{ m{3cm}  m{2.9cm}  m{8.2cm}  }
     \toprule
      \centering Illustration & Regularizer $R(\boldsymbol{\phi})$ &  Description \\ 
      \toprule
    \includegraphics[width=\linewidth]{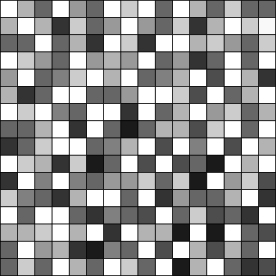}
      & 
\centering $-$
      & 
       \textbf{\textit{No regularization}}:  If regularization is not included in the CE values, the optimization process can converge to non-implementable values, for example, conventional mirror-based spatial light modulators are capable just of emulating binary values passage or blockage of light~\cite{arce2014compressive}.
      \\ 
    \includegraphics[width=\linewidth]{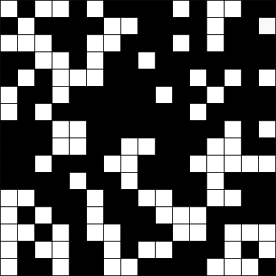}
      & 
	\centering	$ \frac{1}{n} \sum_{{l=1}}^{n} ({\boldsymbol{\phi}_{{l}}})^2({\boldsymbol{\phi}_{{l}}}-1)^2$
      & 
        \textbf{\textit{Binary regularization}}: When this regularization is included the elements of the CE are either (0) or (1), i.e., including this regularization function in the training process encourages that the values of the CE become binary since the minimum of the function is obtained when the entries are 0 or 1~\cite{bacca2020coupled}.
      \\ 
    \includegraphics[width=\linewidth]{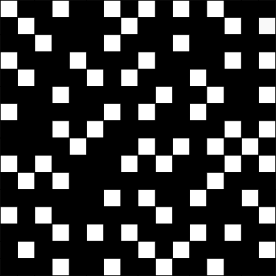}
      & 
\centering	$\frac{\sum_{l=1}^n \left(\prod_{j=1}^S \boldsymbol{\phi}_{l}^j \right)}{n}$
      & 
        \textbf{\textit{Binary regularization + correlation}}: The correlation among the CE $\{\boldsymbol{\phi}^j\}_{j=1}^{S}$ used in a multishot acquisition scheme is crucial to increasing the underlying scene’s captured information, with $S$ as the number of captured snapshots. \textcolor{black}{Specifically, ~\cite{hinojosa2018coded} demonstrated that the reconstruction is improved by reducing the correlation of the CEs at each snapshot.} This desired behavior can be considered in the E2E training by adding a function that minimizes the correlation among the $S$ designed CE. 
      \\ 
    \includegraphics[width=\linewidth]{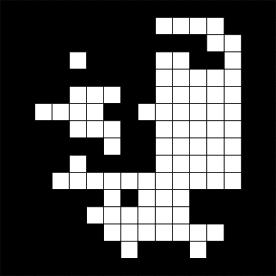}
      & 
       \centering     $  \left(\frac{\sum_{l=1}^n\boldsymbol{\phi}_{l}}{n} - T_r\right)^2$
      & 
        \textbf{\textit{Binary regularization + transmittance}}: To achieve an exact number of non-zero values in the CE, known as transmittance, we can include this regularization function to obtain this desired property, where $Tr\in [0,1]$ is a customizable hyperparameter that denotes the target transmittance level. Remarking that the transmittance level is a crucial property that affects the calibration of the optical coding system and determines the proper utilization of the light in the acquired measurements~\cite{bacca2021deep}.  
      \\ 
    \includegraphics[width=\linewidth]{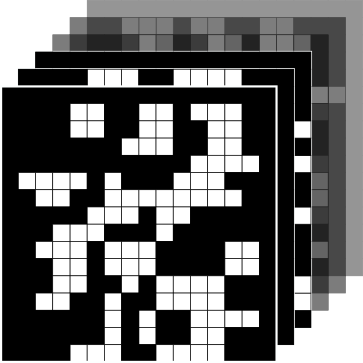}
      & 
\centering	 $ \sum_{j=1}^{S} \sqrt{\sum_{l=1}^n \left( \boldsymbol{\phi}_{l}^j\right)^2}$
      & 
    \textbf{\textit{Binary regularization + Number of shots}}: The aim of acquiring multiple snapshots is to increase the amount of acquired information related to the properties of the scene, improving the task performance. However, the number of snapshots implies a trade-off between the task performance and the needed time for acquiring and processing more snapshots. Therefore, determining the least amount of optimal snapshots $S$ that achieve the highest performance is essential, where the proposed regularizer encourages CE in all zero, which means non-sensed measurements~\cite{bacca2021deep}. 
      \\ 
      \bottomrule
      \end{tabular}
      \end{center}\vspace{-3em}
\end{table}
The regularization over the optical parameters $R_\rho(\boldsymbol{\phi})$ plays a different role than the network weights since \textcolor{black}{they directly change the values of the physical optics}. Therefore, it is useful to promote desired properties of the CE. The main idea of including the regularization in the training is that the gradient of the \textcolor{black}{composed} loss \textcolor{black}{function $\mathcal{L}$} concerning $\boldsymbol{\phi}$ is calculated using the chain rule as:
\begin{equation}
\frac{\partial \mathcal{L}}{\partial \boldsymbol{\phi}} = \frac{\partial \mathcal{L}_{task}}{\partial \mathcal{N}_{\theta}}\frac{\partial \mathcal{N}_{\theta}}{\partial \mathbf{g}} \frac{\partial \mathbf{g}}{\partial \boldsymbol{\phi}} + \rho\frac{\partial R_\rho}{\partial \boldsymbol{\phi}}.
\end{equation}
Therefore, the design of the optical elements is directly influenced by the loss of the task and the regularization function. For example, physical restrictions of the CE implementation process that are not addressed in the parametrization of the optical elements impose constraints on the optimization, such as the CAs entries that must be binary. This constraint can be adequately addressed by including a regularization function. Some examples of regularizer used in the state-of-the-art are presented in Table \ref{tbl:Regularizations}.
Additionally, the parameter $\rho$ plays an essential trade-off role in the optimal performance and the desired properties \edwin{imposed by the regularization}. Therefore, work in~\cite{bacca2021deep} uses an exponential increase strategy, in which the derivative of the loss gives in the first epochs the direction to converge to the desired task values, and then $\rho$ is increased to guarantee the regularization performance.

\begin{figure}[t!]
\centering
\includegraphics[width=0.95\textwidth]{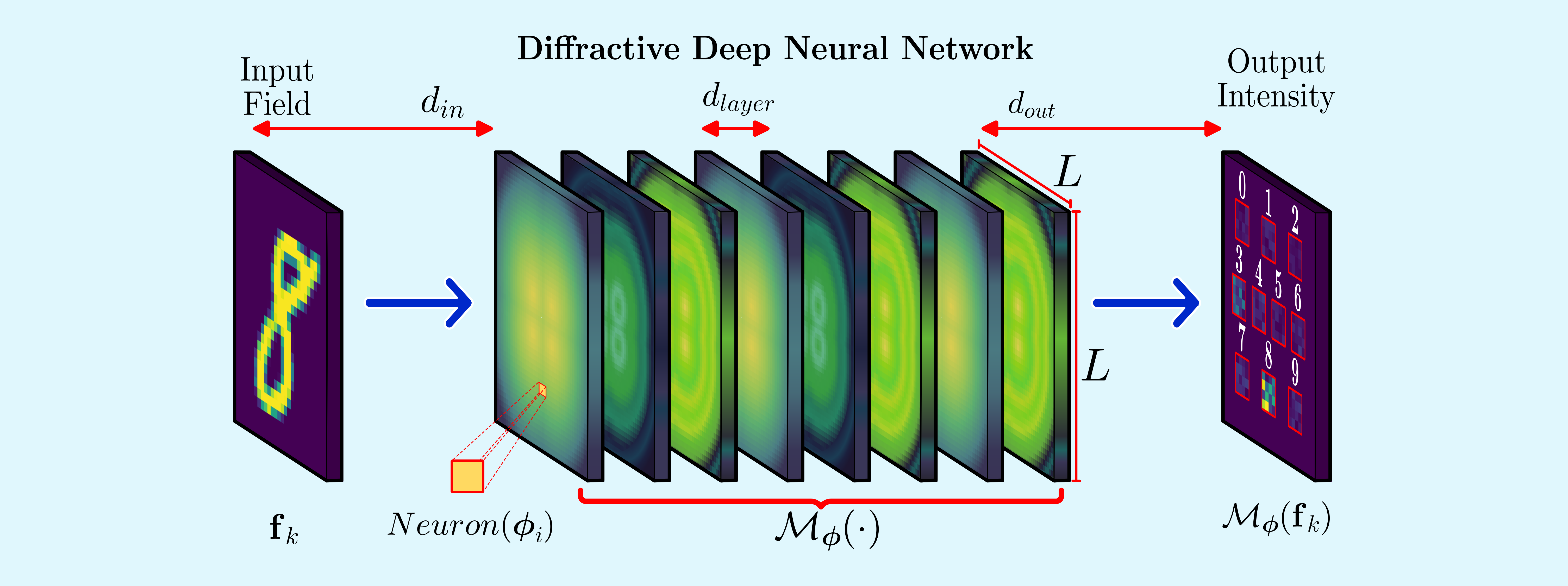}\vspace{-1em}
\caption{A D2NN that consists of $L \times L$ sized \textcolor{black}{CEs} spaced at $d_{layer}$ distance from each other. \textcolor{black}{Each optical layer} consists of neurons which are characterized by a transmission coefficient \textcolor{black}{(neuron)}. The input plane is $d_{in}$ distance away from the first layer of the network, while the output intensity is captured by the detector which is placed $d_{out}$ distance from the last layer of the network. The D2NN depicted here is trained for a classification task. Different applications of a D2NN are discussed in section \ref{subsec:applications_d2nn}.} 
\label{fig:D2NN}\vspace{-2em}
\end{figure}

\section{All Optical Encoding with Diffractive Deep Neural Networks}

Instead of having an encoder consisting of optical layers coupled with an electronic decoder, as discussed in the previous section, a D2NN entirely consists of optical layers that can be trained to learn an arbitrary linear function which can then be performed through a physical setup. Even though the inference and prediction of the physical network is all-optical, learning the parameters to design the network happens through software simulations. As shown in Fig.~\ref{fig:D2NN}, the network consists of several CE, \Ramith{where each CE pixel is called as a neuron} that can be modeled to transmit or reflect an incoming wave~\cite{Lin2018}. Such neurons are connected to other neurons of the following layers through optical diffraction~\cite{Lin2018}. This enables the entire network to be modeled as a differentiable function. Analogous to a standard DNN, we can consider the complex-valued transmission or reflection coefficient of each neuron as a multiplicative \emph{bias} term, where both the amplitude and phase can be treated as learnable parameters. They can be adjusted iteratively via an error back-propagation method. As mentioned in \cite{Lin2018}, the input object information can be encoded in the amplitude and/or phase channel of the input field, and the desired result will be captured as the intensity of the output field by a detector. The modelled optical system ($\mathcal{M}_{\boldsymbol{\phi}}$) is trained to optimize the parameters, i.e.,
\begin{equation}
\label{eq:E2E_d2nn}
\hspace{-0em}\boldsymbol{\phi}^* \hspace{-0em}=\hspace{-0em}\argmin_{\boldsymbol{\phi}} \sum_{k}^{K}\hspace{-0em} \mathcal{L}_
		{task}\left( \mathcal{M}_{\boldsymbol{\phi}}(\mathbf{f}_k) , \mathbf{d}_k\right),\hspace{-0em}
	\end{equation}
where $\boldsymbol{\phi}^*,\, \mathbf{f}_{k},\, \mathbf{d}_k$ denote the optimal transmission coefficient parameters, input field and target from the training database consisting of $K$ instances, respectively. Once the network is trained, the design of the D2NN is fixed and the layers can be fabricated subjected to the constraints as discussed in section \ref{subsec:realizaton_complexities_d2nn}. The network can then be used to perform the learned task at the speed of light with little or no power consumption.


During the simulation of the forward propagation of light through a D2NN, we may directly use the RS integral~\eqref{eq:rayleigh_sommerfeld} and~\eqref{eq:direct1} to compute the field for each neuron in all the layers. This approach is denoted as the \emph{direct method} and can be implemented to perform computations in parallel. However, the direct method requires $N^2$ number of $N \times N$ matrices for each layer, \Ramith{making the memory requirement for computations at each layer $N^4$}. This requirement grows extensively large \textcolor{black}{as} the number of neurons \textcolor{black}{increases resulting} in large computational time requirements.


\label{sec:all_optical}
%
\Ramith{In order to eliminate this drawback, the AS method can be employed to efficiently compute the fields of neurons in a D2NN.}
The electromagnetic light-wave field and the AS can be expressed as a Fourier relationship using the discrete Fourier transform (DFT) model based on the fast Fourier transform (FFT) algorithm. The DFT-based approximation is commonly used in the D2NN approaches to overcome the intrinsic high computational complexity of the direct model. For instance, consider two CEs located at the $z_{l-1}$ and $z_{l}$ planes in the   D2NN-based COI system  as shown in Fig. ~\ref{fig:D2NN}. In this case, the CE's width and height with $L$-size are sampled as $N\times N$ with $N = L/dx$ where $\boldsymbol{\phi}^{l-1}$ and $\boldsymbol{\phi}^l$ are \Ramith{the complex transmission coefficients of the optical layers}, respectively.

\begin{figure}[t!]
    \centering
    \includegraphics[width=0.92\linewidth]{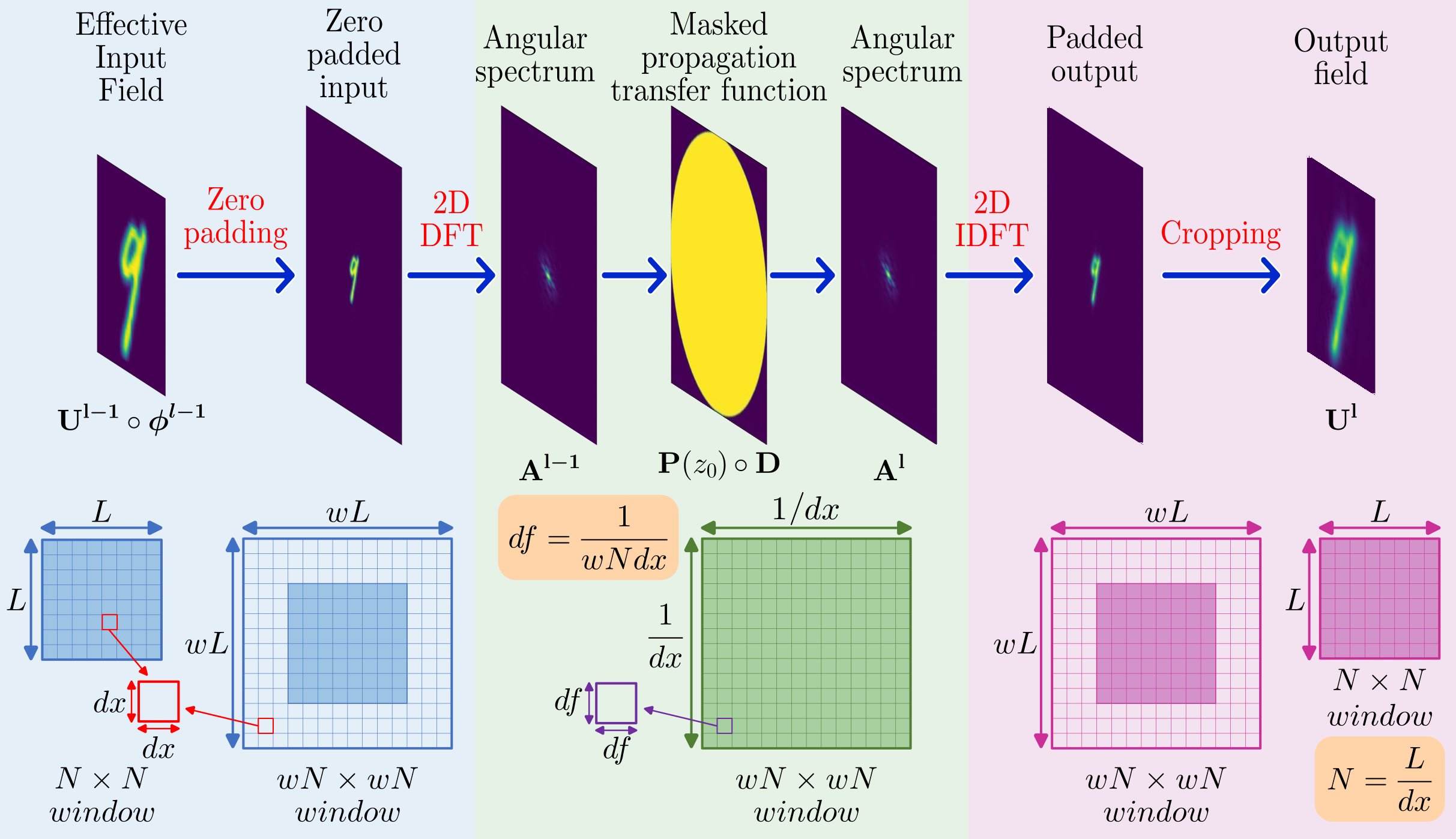}\vspace{-1.5em}
    \caption{Computational pipeline of the AS method. The 2D-DFT operation is performed after a DFT shift operation to make sure that the origin of the AS aligns with that of the propagation transfer function. In the figure $\circ$ denotes the element-wise matrix multiplication.}\vspace{-2em}
    \label{fig:fft-as_pipe}
\end{figure}

Fig.~\ref{fig:fft-as_pipe} shows the computation pipeline of the output field using the AS method. Initially, the input field is sampled with a sampling interval of $dx$ in the spatial domain.  \bacca{According with the modeling explained in section~\ref{section:2}, the propagation and modulation of the input filed, i.e., the $(l-1)^{\textrm{th}}$ optical layer, is given by, \Ramith{$\mathbf{U}^{l-1} \circ \boldsymbol{\phi}^{l-1}$} where $\circ$ is the element-wise matrix multiplication. } \textcolor{black}{Then, the 2D-DFT of a zero padded version of the encoded field is applied to obtain the AS given by $\mathbf{A}^{l-1}$ with a  computational window of the size $wN \times wN$. } 
The propagation transfer function matrix $\mathbf{P(z_0)}$ at $z_0$ is also created in the computation window with a similar shape as $\mathbf{A}^{l-1}$. This matrix is the discrete representation of the propagation transfer function given in~\eqref{eq:prop_tf}. The region where $f^2_x + f^2_y > 1/\lambda$ in $\mathbf{P(z_0)}$ corresponds to evanescent waves which do not propagate energy along the $z$-axis. Hence, they are filtered out using a \textcolor{black}{binary} mask $\mathbf{D}$ as 
is shown in Fig.~\ref{fig:fft-as_pipe}.
The AS of the field at $z_0$ is obtained by the element-wise multiplication of $\mathbf{A}^{l-1}$ and the masked propagation transfer function. Then 
a 2D-IDFT operation \textcolor{black}{is performed} to obtain the padded resulting field. Finally, the padding is removed to retrieve the resulting field $\mathbf{U}^l$. Since all the computations happen in the computational window, the memory requirement for the computations at each layer of the AS method is $(wN)^2$. Therefore, a \emph{significant reduction} in the memory requirement can be achieved with the AS method compared to the direct method for typical D2NNs. For an example, the classifier D2NN used in~\cite{Lin2018} has $200 \times 200$ neurons per layer ($N=200$). If the direct method is used for the implementation, it requires matrices with a total number of $1.6$ billion ($=200^4$) elements. In contrast using the AS method with $w=4$, which is sufficient to provide accurate results, each layer only requires matrices with a total number of $640,000=(4 \times 200)^2$ elements. This is a $2500$ times reduction in the memory requirement per single layer.

\section{Real Implementations}

 \subsection{Implementation and Fabrication of Optical Coding Elements}
\label{subsec:realizaton_complexities_d2nn}

When the  E2E-based CE design ends, they are manufactured or implemented following the learned  parameters. The digital micromirror device (DMD)~\cite{arce2014compressive}, or Liquid crystal on silicon (LCoS)\cite{hinojosa2021learning} optical elements are commonly used to mimic and validate the trained CE's pattern. DMD-based CE experiments exploit the spatial modulation capabilities of the $M\times N$ array of metalized polymer mirrors bonded to a silicon circuit. Each micromirror can have a pixel size resolution of $~$1-25 $\mu$m with two degrees of freedom [i.e., rotation around two orthogonal axes] and refreshing rates of up to 25 kHz. The DMD adjustable angle property is used to let pass or block the light, which allows it to work as a binary coded aperture. In this path, DMD-based gray-scale modulation approaches have been proposed by synchronizing the DMD with the sensor's integration time ~\cite{bacca2021deep}, i.e., several DMD patterns are used for generating \textcolor{black}{the coded measurements}. In contrast, the LCoS devices comprise a layer of liquid crystal sandwiched between a top sheet of glass coated with a transparent electrode and a pixelated silicon substrate. Because the silicon is reflective, it serves as a mirror element for each pixel, with the strength of the reflection electronically controlled by the amount of light transmitted through the above liquid crystal. The LCoS capability to regulate light transmission has constituted it as the most used optical device to modulate the wavefront's phase.

\Ramith{The construction of a CE with a fixed pattern} allows considerably reducing its size as well as its cost and improves its portability. For instance, a cheap way to perform CA is through photography-film. Accordingly, work in \cite{arguello2021shift} prints a color-CA using a FUJICHROME Velvia 50 transparency film (35 mm, 36 exposures), with an ISO of 50, which provides high sharpness and daylight-balanced color. 
On the other hand, \Ramith{the CEs used in the D2NN~\cite{Lin2018} were fabricated to work in Terahertz (THz) wavelengths.} This work realized D2NNs which consist of $300$ $\mu$m$\times 300$ $\mu$m$\textrm{/}400$ $\mu$m$\times 400$ $\mu$m sized neurons with millimeter/centimeter-level layer distances. The operating wavelength was $749.48$ $\mu$m (THz range). But for imaging, visible light is the most prominent operating wavelength range. To scale down the D2NN in THz proposed by \cite{Lin2018} to visible wavelength (e.g. $632.8$ nm), nanometer-scale optical neurons are required in contrast to the millimeter-scale used in THz. Fabricating such features having feature sizes smaller than the diffraction limit of visible light is challenging, and therefore requires special fabrication methods. 
     \textcolor{black}{In the following we briefly present current} potential
fabrication technologies \textcolor{black}{including} electron beam lithography (EBL), two-photon polymerization, and implosion fabrication.

\textcolor{black}{EBL is a technique with an extremely high diffraction-limited resolution (compared with conventional optical or ultraviolet photolithography) that transfers desired patterns having  feature sizes around 10 nm onto an EBL-resist. Then, the carved resist is dissolved in a chemical bath and filled with the desired material.} Finally, the resist is dissolved \textcolor{black}{again} to obtain the material with the desired patterns. Two-photon polymerization is a \textcolor{black}{non-linear} laser writing technique \textcolor{black}{based on the simultaneous absorption of two photons in  a photosensitive material, e.g., polymers. This technique allows the creation of complex three-dimensional structures with feature sizes on the order of 100 nm.}
Implosion fabrication \textcolor{black}{is a method that creates large-scale objects embedded in expanded hydrogels and then shrinks them to nanoscale. Specifically, this technique uses an absorbent material made of polyacrylate as the scaffold for their nanofabrication process. The scaffold is bathed in a solution containing fluorescein molecules, which attach to the scaffold in defined three dimensional patterns when they are activated by laser light.} The patterned and functionalized gel scaffolds are shrunken using acid/divalent cations. Then, the final structure is obtained after a dehydration process.\vspace{-1em}

\subsection{Fine-tuning of Optical Coding Elements}
\begin{figure}[!t]
	\centering
	\includegraphics[width=\textwidth]{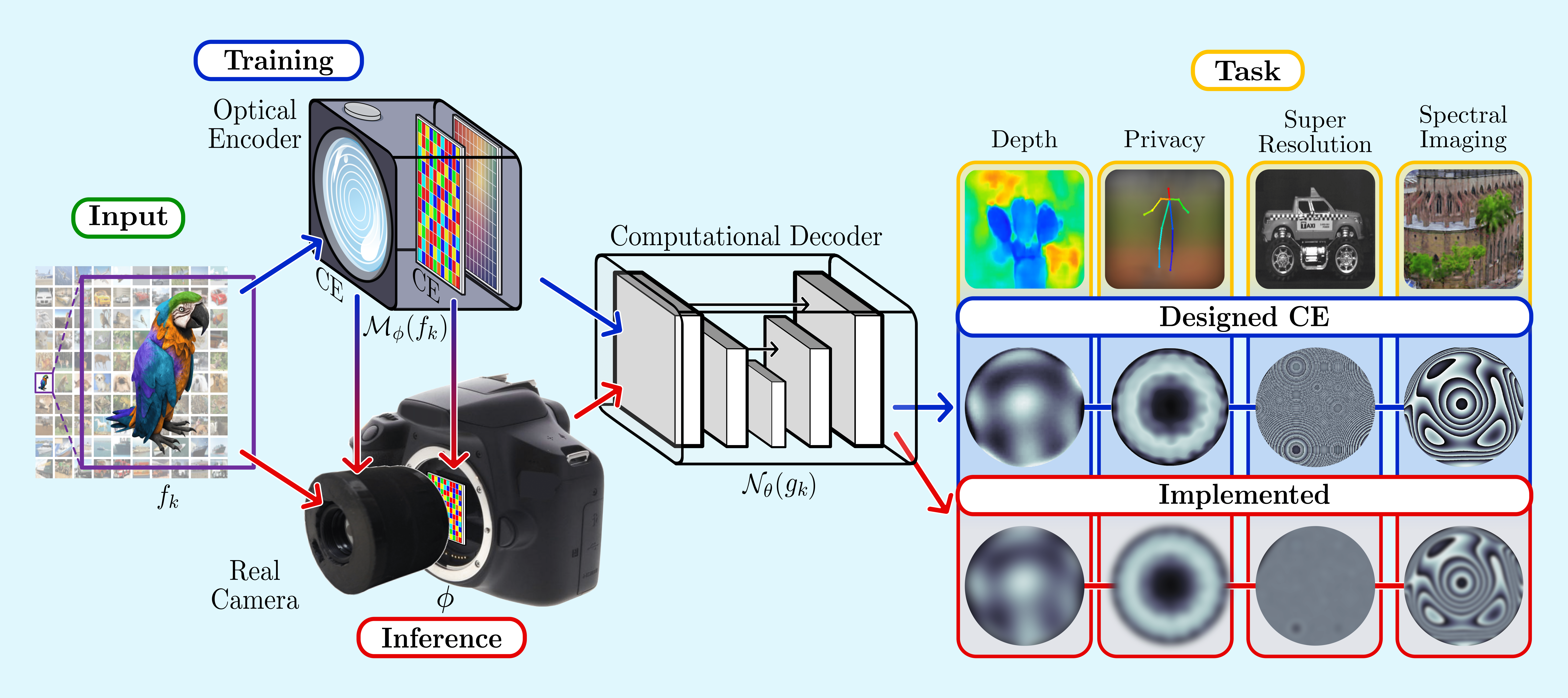}\vspace{-1.5em}
	\caption{Illustration of the real implementation using the deep optical coding design. With the real optical system, new measurements are acquired and the same trained network is used to perform the desired task (Depth~\cite{sitzmann2018end}, Privacy \cite{hinojosa2021learning}, Super Resolution~\cite{sitzmann2018end}, Spectral Imaging~\cite{arguello2021shift}) directly to the measurements. \vspace{-2em}
	}\label{Real_imp}
\end{figure}

 One of the essential steps in E2E-based CE designing is the DNN experimental validation, i.e., with measurements acquired by a test-bed system. This validation is conducted by first assembling the COI system using the CE obtained from the E2E optimization strategy, followed by the characterization of the optical system response, \textcolor{black}{as illustrated in Fig~\ref{Real_imp}}. Then, a DNN re-training, termed as fine-tuning \cite{arguello2021shift}, is carried out by \textcolor{black}{decoder weights, while optical layer weights are fixed with the characterized CE values by passing the training dataset through the real sensing model.  For example, in \cite{arguello2021shift}, this process was carried out by obtaining the PSFs of the implemented system and a small set of 10 ground-truth spectral scenes and \Hans{their} corresponding measurements to retrain the decoder network along $100$ epochs using a small learning rate of $10^{-6}$.} This process is indispensable to improve the decoder robustness and fidelity in real scenarios since the propagation analytical model does not consider the non-ideal CE characteristics, such as fabrication errors or COI system misalignments as illustrated in Fig. \ref{Real_imp}. For instance, in \cite{arce2014compressive}, the designed pixelated optical filters are not perfectly square as expected, or in  \cite{hinojosa2021learning}, the resulting experimental point spread function has slight distortions associated with objective and relay lenses imperfections. After calibrating the camera and re-training the decoder, the task-oriented COI system is ready. A remarkable feature of the task-oriented CE designing framework resides in optimally adapting the CE distribution according to the COI system configuration and the task specifications.  For instance, task-oriented DOE designing has been studied in several COI applications such as, privacy preserving pose estimation~\cite{hinojosa2021learning}, video recovery  \cite{vargas2021time}, super-resolution \cite{sitzmann2018end}, depth estimation \cite{sitzmann2018end}, and  spectral imaging~\cite{arguello2021shift}, as is illustrated in Fig. \ref{Real_imp} where it is highlighted that each task generates a unique DOE structure.

\section{Computational Applications}

The E2E optimization framework has been applied in diverse applications such as reconstruction, classification or object detection. This survey presents some relevant applications that have leveraged E2E optimization and describe the specific encoder-decoder parameters as well as the cost functions employed to optimize them.

\subsection{Spectral Imaging}

Spectral imaging aims to capture 2D images of the same scene at different wavelengths. Spectral images enable a diverse range of applications, including medical imaging, remote sensing, defense and surveillance, and food quality assessment~\cite{arce2014compressive}. The amount of spatial information across a multitude of wavelengths represents one of the main challenges of the traditional scanning-acquisition imaging systems since to obtain several high-definition images, these systems require a long exposure time, therefore limiting their use in real-time applications~\cite{arce2014compressive}. To overcome this limitation, several computational approaches have been proposed to acquire a single snapshot and recover the desired spectral cube in a computational stage \cite{arce2014compressive}. More recent approaches employed the E2E framework to optimize the CEs jointly, and the image processing algorithm leading to state-of-the-art results~\cite{arguello2021shift,vargas2021time}.

\begin{figure}[!t]
	\centering
	\includegraphics[width=\textwidth]{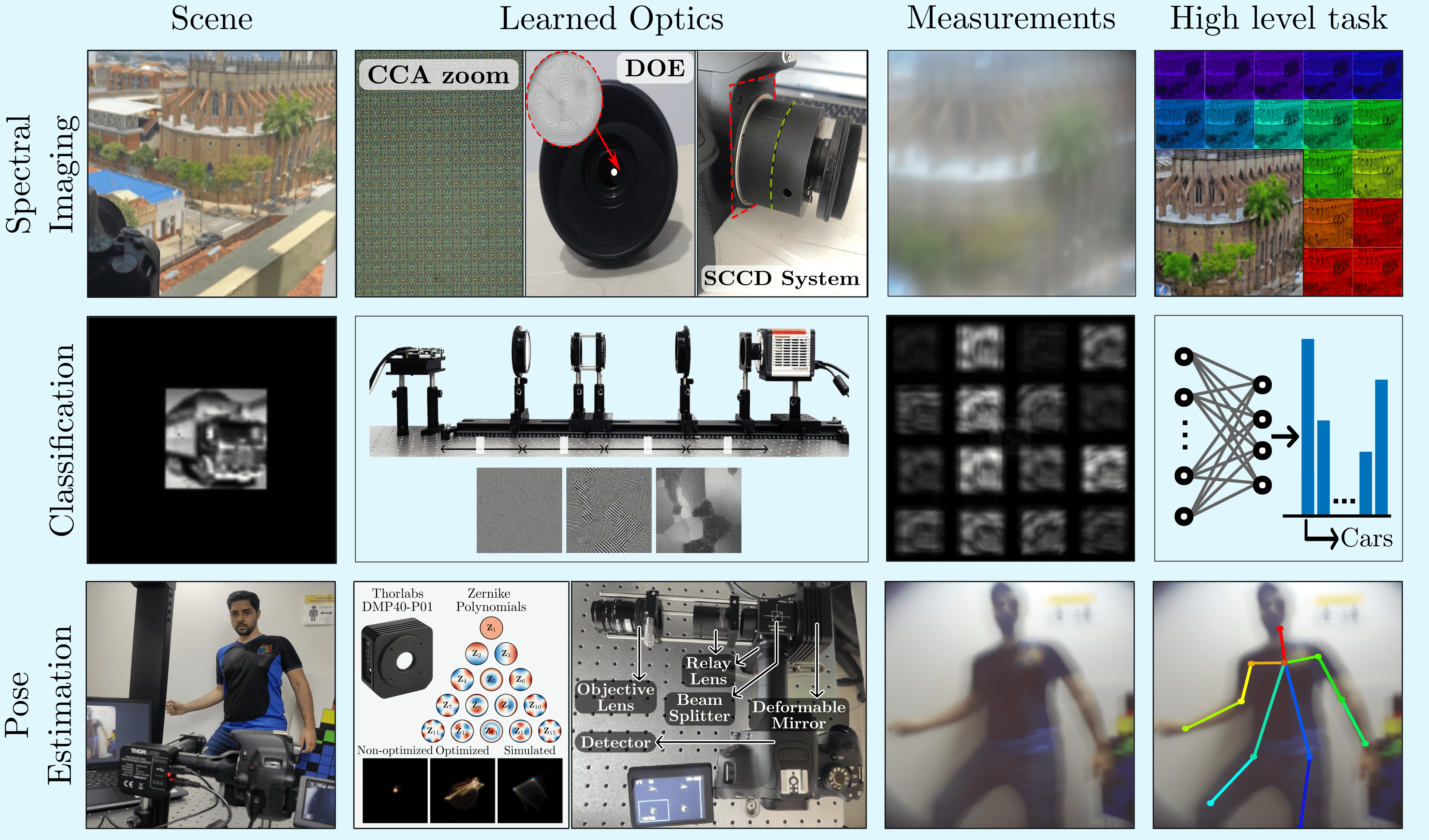}\vspace{-1.5em}
	\caption{Three representative optical encoding and computational decoding applications that employ E2E optimization,  (Top) spectral imaging, (Middle) image classification, and (Bottom) human pose estimation.}\vspace{-2em}
	\label{fig:All_aplications}
\end{figure}

For instance, \Miguel{the authors in \cite{arguello2021shift} propose an optical setup} composed of a colored CA and a DOE for snapshot spectral imaging, as shown at first row of Fig. \ref{fig:All_aplications}. The proposed optical system has a different PSF for each pixel, \textcolor{black}{improving the coding of the optical system}. In this case, the optical parameters $\boldsymbol{\phi}$ consist of a binary spatial mask that selects a given primary color for each spatial pixel location of the CCA and a set of coefficients that weight the Zernike polynomials to represent the DOE height map \cite{sitzmann2018end}. As an electronic decoder, a deep network based on UNet with skip connection was employed \cite{bacca2021deep}. The decoder parameters $\boldsymbol{\theta}$ are the weights of the UNet network. To optimize $\boldsymbol{\phi}$ and $\boldsymbol{\theta},$ the mean square error (MSE) metric was employed setting $\mathcal{L}_{task}$ in the Eq. \eqref{eq:E2E_general} to $\mathcal{L}_{task}(\mathbf{x},\hat{\mathbf{x}})= \|\mathbf{x}-\hat{\mathbf{x}}\|_2^2$. Furthermore, the binary regularization introduced in \eqref{tbl:Regularizations} is employed to promote binary values on the spatial mask selecting the optimal color.



\subsection{Image Classification}
Image classification is a complex task that has been addressed successfully with deep convolutional neural networks (CNNs). However, the deployment of CNN in mobile or low-resource devices is prohibitive \edwin{due to} the high costs of memory, processing, and power. To increase \edwin{computational} efficiency, recent approaches design CEs that can perform computations via light propagation and can work as a co-processing unit. Thus, the image information can be processed at the speed of light with reduced power consumption. For instance, the authors in \cite{Chang2018HybridClassification} proposed a hybrid optical-electronic CNN that implements the first layer of a CNN using an optimizable DOE and the remaining layers in an electronic decoder as is illustrated in the middle row of  Fig~\ref{fig:All_aplications}. The optical parameter in this system is the height profile of a DOE that produces the kernels of the first optical layer. The digital decoder is just a simple non-linear activation function followed by a fully connected layer. Similar to standard optimization methods for classification, the cross-entropy loss is employed to optimize the kernels generated by the DOE and the fully connected linear layer of the decoder. Since the generated PSFs from the DOE cannot have negative values, an additional non-negative regularization is employed.


\subsection{Privacy Preserving for Human Pose Estimation}
Cameras in smartphones, cars, homes, cities collect a huge amount of information in an always-connected digital world. However, this raises a big challenge in the preservation of privacy. Privacy-preserving approaches rely on a traditional digital imaging system.  For example, the detection of privacy-sensitive everyday situations can be done by software. Then an eye tracker’s first-person camera can be enabled or disabled using a  mechanical shutter \cite{hinojosa2021learning}.  However,  such a  method performs software-level processing on videos acquired by traditional cameras, which may already contain privacy-sensitive data that could be exposed in an attack. A recent approach based on the design of CE proposes designing a freeform lens that protects sensitive data while maintaining useful features for perceiving humans in the scene, specifically, human pose estimation (HPE) using hardware-level protection~\cite{hinojosa2021learning} as is shown in the bottom of Fig~\ref{fig:All_aplications}. Similar to previously presented applications, in this work, the optimization of the optical encoder is carried out through the varying surface height of a lens parameterized using the Zernike polynomials. \Ramith{To obtain the locations of body keypoints}, the OpenPose network composed of a VGG backbone and two CNN branches is employed. The cost function $\mathcal{L}$ of the proposed privacy HPE framework consists of the OpenPose loss $\mathcal{L}_p$ plus a cost function that enforces image degradation. The latter term, in this case, consists of the negative of the MSE between the output private image of the designed camera and the high-resolution latent image.
\subsection{All-Optical Applications of D2NNs}
\label{subsec:applications_d2nn}
\begin{figure}[!t]
     \centering
     \includegraphics[width=1\textwidth]{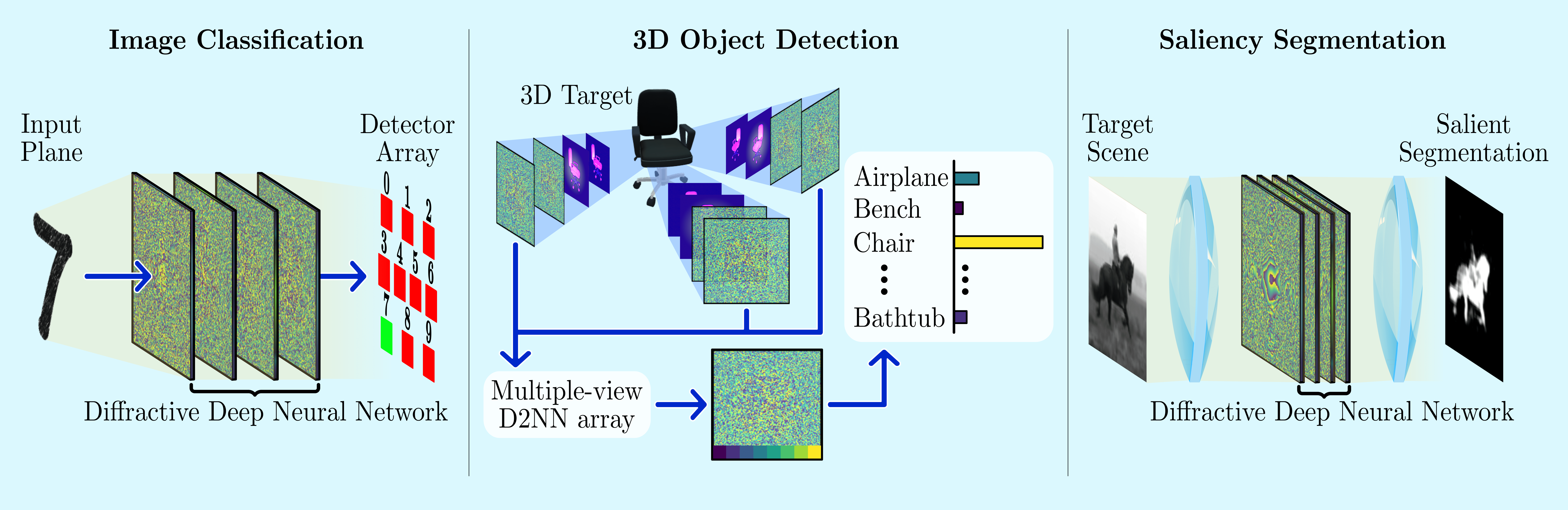}\vspace{-1.5em}
     \caption{Three types of all-optical applications of D2NNs for image classification (left), 3D object detection using multi-view D2NNs (middle), saliency segmentation by placing the D2NN at the Fourier plane of the 4-$f$ system (right).}\vspace{-2em}
     \label{fig:d2nn_application}
 \end{figure}
In contrast to optoelectronic-based applications, D2NNs provide an opportunity to perform \textit{all-optical} systems that can perform computations at the speed of light through optical diffraction. As the layers do not require power to operate, the inference of these networks becomes scalable with an increasing number of neurons. The first demonstration of D2NNs presented experiments in image classification, and amplitude-to-amplitude imaging \cite{Lin2018}. For experimental evaluation of image classification, \Ramith{$5$-layer D2NNs have been trained} with trainable phase-only transmission coefficients ($\boldsymbol{\phi}$) while using MSE as the loss function ($\mathcal{L}_{task}$) to quantify classification accuracy for the MNIST and FashionMNIST datasets separately. Furthermore, the D2NN was physically implemented with 3D printed layers, each having $200 \times 200$ neurons in $8$ cm $\times$ $8$ cm area, and \Ramith{the frequency of the illuminating wave was $0.4$ THz}. The all-optical classifiers reported $88\%$ and $90\%$ accuracy for classifying 3D printed handwritten digits and fashion products, respectively.


Several research works have extended D2NNs for tasks such as 3D object recognition \cite{Shi2021MultipleviewRecognition} and saliency segmentation \cite{Yan2019Fourier-spaceNetwork}. To address the task of 3D object recognition, multiple D2NNs have been employed to capture light fields from different views and aggregate them together to obtain better accuracy compared to a single view. \Ramith{Each of these D2NNs is composed} of two layers and learns to produce an optical spot pattern from which a one-hot encoding can be obtained. These networks employ softmax cross-entropy as the training loss function. In inference, a weighted summation of the sub-light fields is obtained to perform the prediction. \Ramith{While the previously discussed research works learn a transformation through a D2NN in the real space, authors in \cite{Yan2019Fourier-spaceNetwork} explore learning such transformation on the Fourier space to perform saliency segmentation.  They utilize a 4-$f$ system ~\cite{Lin2018} which is an optical system with two lenses placed 2 focal lengths ($f$) apart from each other that perform a cascade of Fourier transforms. They place a D2NN in the Fourier plane (i.e. one focal length after the first lens) of the 4-$f$ system and perform simulations} to show that the saliency map can be obtained all optically by training to minimize the MSE between the output intensity distribution and the ground truth saliency map. We present a summary of these all-optical applications of D2NNs in Fig.~\ref{fig:d2nn_application}.

\section{Conclusions and future directions}
This paper summarized the essential concepts for data-driven CE design via \textcolor{black}{the joint optimization of optics and the image processing algorithm. These concepts include} CE parametrization, modeling physics-based propagation of light, and the underlying optimization framework \textcolor{black}{and are} illustrated via \textcolor{black}{the design of} two of the most popular customizable optical elements, CA and DOE. We showcased practical applications of this rapidly developing framework in various imaging problems such as spectral imaging, privacy-preserving for human-pose estimation, and image classification. Furthermore, we illustrate the data-driven CE design in an all-optics framework that performs DNN inference at the speed of light. 

Although optical coding design has become the state-of-the-art way to design the optical coding elements in multiple COI applications, it is still evolving. In this direction, there are some challenges and open problems when applying the E2E framework. For instance, what is the proper training scheme? The current optimization framework mainly lies in the backpropagation algorithm. Therefore, the optical layer may not be properly optimized for being naturally the first layer of the coupled deep model and thus be fundamentally limited by the gradient vanishing of the CE (i.e., gradients converge to zero). Another essential question is the influence of proper initialization of the optical parameters well-studied for DNN layers but not for optical layers. \textcolor{black}{Since the optical design is obtained from a database, the results lack interpretability. Some works try to analyze the PSFs of the system, for instance, by measuring the coherence between bands\cite{arguello2021shift} or giving some intuition about the results~\cite{sitzmann2018end} based on the concept of PSF engineering. However, designing metrics or provide some guidance to understand the results is still an open problem.} In this reasoning, even though custom neural networks are suitable digital decoders, there are no clear insights related to the characteristics of deep architectures that obtain the best deep optical design. Therefore, several issues related to the interaction between the optical encoder and digital decoder are still to be understood. 

From the perspective of constructing the optimized physical imaging system, there is also a fundamental issue regarding the mismatch between the mathematical and the real model. This problem is addressed by calibrating the assembled optical system and/or \textcolor{black}{fine-tuning} the digital decoder. Additionally, the D2NNs only perform linear operations in the all-optics framework. Thus, there still exist limitations in the optical implementation of inherent non-linearities such as activation functions and pooling operations. To overcome this limitation, a few recent works are based on mode coupling in a multi-mode optical fiber \cite{Tegin2021ScalableOperator} or based on nonlinear material light interactions such as electromagnetically induced transparency and saturable absorption \cite{Lin2018}. However, making an all-optical model that considers the latest ideas in DNNs is still an open field.

 \bacca{Finally, The CE designing framework has focused on a limited set of elements relegating the other elements and parameters that compose the COI systems to a pre-designed structure. This fixed structure suggests a future direction to explore more flexible camera designs. \textcolor{black}{For instance, physical considerations that remain fixed until now, such as the distance between the optical elements, the sensor size and the number of measurements, can be considered in the deep optical design with proper modeling}. We anticipate that in the near future, \textcolor{black}{we will use the compact cameras in our smartphones} or devices designed using this methodology for diverse applications, mainly in specific task-driven software.}

\tiny{\bibliography{sample,references}}
\tiny{\bibliographystyle{ieeetr}}
\vspace{-9em}
\begin{IEEEbiographynophoto}{Henry Arguello} (henarfu@uis.edu.co) received the Master's degree in electrical engineering from the Universidad Industrial de Santander, Colombia, in 2003, and the Ph.D. degree from the Electrical and Computer Engineering Department, University of Delaware in 2013. In 2020 he was a visitor professor with the Computational Imaging group in Stanford. He is currently a Titular Professor with the Systems Engineering Department, Universidad Industrial de Santander, Bucaramanga, Colombia. He is a senior member of IEEE. He is the president of the signal processing chapter, Colombia. His current research interests include computational imaging techniques, high dimensional signal coding and processing, and optical design.
\end{IEEEbiographynophoto}
\vspace{-9em}
\begin{IEEEbiographynophoto}{Jorge Bacca} (jorge.bacca1@correo.uis.edu.co) received the B.S. degree in computer science and the Ph.D. degree in Computer Science from the Universidad Industrial de Santander (UIS), Bucaramanga, Colombia, in 2017 and 2021, respectively. He is currently a Postdoctoral Researcher at UIS. He is a consulting associate editor for the IEEE Open Journal of Signal Processing. He is an author of more than 10 journal papers and 30 proceedings conferences related to optical design and high-level tasks directly to raw optical measurements. His current research interests include inverse problems, deep learning methods, optical imagining, and hyperspectral imaging.
\end{IEEEbiographynophoto}\vspace{-9em}

\begin{IEEEbiographynophoto}{Hasindu Kariyawasam} (170287A@uom.lk) completed his B.Sc. degree in electronics and telecommunication engineering at the University of Moratuwa, Sri Lanka. He is currently a post-baccalaureate research fellow in Center for Advanced Imaging at Harvard University. He has also worked as a visiting researcher (student) at the University of Sydney, Australia. His research interests include physics based machine learning, digital system designing, optical neural networks, human computer interaction, signal processing and natural language processing.
\end{IEEEbiographynophoto}\vspace{-9em}

\begin{IEEEbiographynophoto}{Edwin Vargas} (edwin.vargas4@correo.uis.edu.co) Master and B.S degree in electronical engineering from the  Universidad  Industrial  de  Santander,  Bucaramanga,  Colombia, in 2018,and 2016, respectively. He is currently a Ph.D. candidate in Engineering at Universidad Industrial de Santander. His research interests includes high-dimensional signal processing, compressive sensing, and the development of new computational cameras using deep learning. 
\end{IEEEbiographynophoto}\vspace{-9em}
\begin{IEEEbiographynophoto}{Miguel Marquez} (hds.marquez@gmail.com) received his B.Sc. degree in Computer Science and his M.Sc. degree in Applied Mathematics from the Universidad Industrial de Santander, Colombia, in 2015 and 2018, respectively. He is currently working toward a Ph.D. degree in Physics with the Universidad Industrial de Santander. In 2021 he was an Intern
with LACI laboratory at the Institut National de la Recherche Scientifique (INRS)–Université du Québec, Canada. His main research interests include computational imaging, compressive sensing, optimization algorithmic, and optical system design.
\end{IEEEbiographynophoto}\vspace{-9em}

\begin{IEEEbiographynophoto}{Ramith Hettiarachchi}(170221T@uom.lk) is currently a final year undergraduate pursuing the B.Sc. Engineering (Honours) degree in Electronics and Telecommunication from University of Moratuwa, Sri Lanka.  He worked as a Student Researcher at the Robotics and Autonomous Systems Group at CSIRO Data 61, Brisbane, Australia. His research interests include optical computing, computer vision, and machine learning.
\end{IEEEbiographynophoto}

\begin{IEEEbiographynophoto}{Hans Garcia}(hans.garcia@saber.uis.edu.co) Master and B.S degree in electronical engineering from the  Universidad  Industrial  de  Santander,  Bucaramanga,  Colombia, in 2018 and 2016, respectively. He is currently a Ph.D. candidate in Engineering at Universidad Industrial de Santander.  His research interests include multi-resolution image recovery, compressive spectral imaging, computational imaging, remote sensing, and compressive sensing, also in the last years he is workings on optics implementation of designed coding imaging systems.
\end{IEEEbiographynophoto}\vspace{-9em}

\begin{IEEEbiographynophoto}{Kithmini Herath}(170213V@uom.lk) completed her Honours Degree of Bachelor of the Science in Engineering at the Department of Electronic and Telecommunication Engineering, University of Moratuwa, Sri Lanka. She is currently a post-baccalaureate research fellow at the Division of Science of the Faculty of Arts and Sciences at Harvard University. She has also worked as a visiting researcher (student) in the School of Computer Science within the Faculty of Engineering at the University of Sydney, Australia. Her research interests include signal processing, machine learning, computer vision and human computer interaction.
\end{IEEEbiographynophoto}\vspace{-9em}

\begin{IEEEbiographynophoto}{Udith Haputhanthri}(170208K@uom.lk) is a final year biomedical engineering undergraduate at the Department of Electronic and Telecommunication Engineering, University of Moratuwa, Sri Lanka. He is currently working as a part-time remote visiting undergraduate research fellow at, Center for Advanced Imaging, Faculty of Arts and Sciences at Harvard University parallel to his university studies. His research interests include computer vision, deep learning, computational imaging, and neuroscience.
\end{IEEEbiographynophoto}\vspace{-9em}

\begin{IEEEbiographynophoto}{Balpreet Singh Ahluwalia} (balpreet.singh.ahluwalia@uit.no) is a professor at the Department of Physics \& Technology, UiT The Arctic University of Norway. He is also affiliated to the Department of Clinical Sciences, Intervention and Technology (CLINTEC), Karolinska Institute in Sweden. Previously, Ahluwalia was a visiting scientist at the Centre of Biophotonics of Science and Technology, University of California Davis (2012) and at the Opto-electronics Research Centre (ORC), University of Southampton (2008, 2019) and invented chip-based optical nanoscopy. He also co-founded Chip NanoImaging AS. Ahluwalia was awarded Tycho J{\ae}ger Prize in Electro-optics and  University's Research \& Development Award in  2018. 
\end{IEEEbiographynophoto}\vspace{-9em}

\begin{IEEEbiographynophoto}{Peter So} (ptso@mit.edu) is a professor in the Department of Mechanical and Biological Engineering in the Massachusetts Institute of Technology.  Prior to joining MIT, he obtained Ph.D. from Princeton University in 1992 and subsequently worked as a postdoctoral associate in the Laboratory for Fluorescence Dynamics in the University of Illinois in Urban-Champaign.  His research focuses on developing high resolution and high information content microscopic imaging instruments.  These instruments are applied in biomedical studies such as the non-invasive optical biopsy of cancer. He is currently the Director of the MIT Laser Biomedical Research Center, a NIH NIBIB P41 research resource.
\end{IEEEbiographynophoto}\vspace{-9em}

\begin{IEEEbiographynophoto}{Dushan N. Wadduwage} (wadduwage@fas.harvard.edu) is a John Harvard Distinguished Science Fellow in Imaging, and the principle investigator of Wadduwage Lab at Harvard's Center for Advanced Imaging. Prior to joining Harvard, Wadduwage obtained his Ph.D. from National University of Singapore under Singapore–MIT Alliance  Fellowship (2016) and subsequently worked as a postdoctoral associate in the MIT Laser Biomedical Research Center. His research focuses on developing learning based differentiable microscopy systems for high-throughput high-content biomedical applications.
\end{IEEEbiographynophoto}\vspace{-9em}

\begin{IEEEbiographynophoto}{Chamira U. S. Edussooriya} (chamira@uom.lk) received the Ph.D. degree in Electrical Engineering from the University of Victoria, Victoria, BC, Canada, in 2015. He has been a Senior Lecturer at the Department of Electronic and Telecommunication Engineering, University of Moratuwa since 2016, and a Courtesy Post-Doctoral Associate at the Department of Electrical and Computer Engineering, Florida International University, Miami, FL, USA since December 2019. His current research interests include analysis and design of low-complexity multi-dimensional digital filters, 4-D light field and 5-D light field video processing, and computational imaging.   
\end{IEEEbiographynophoto}

\end{document}